\documentclass[journal]{IEEEtran}

\usepackage{hyperref}

\ifCLASSINFOpdf
\else
\fi

\usepackage{amsmath}
\usepackage{graphicx}
\usepackage{multirow}
\usepackage{subfigure}
\usepackage{float}
\usepackage{epstopdf}
\usepackage{graphicx}
\usepackage{url}
\renewcommand{\vec}{\boldsymbol} 

\usepackage{color}

\begin{document}

\title{Audio-Visual Kinship Verification}


\author{Xiaoting~Wu, Eric~Granger,~\IEEEmembership{Member,~IEEE},~Xiaoyi~Feng
\thanks{This work is supported in part by the China Scholarship Council (grant 201706290103), the Academy of Finland, and the Natural Sciences and Engineering Research Council of Canada.}
\thanks{X. Wu is with the Center for Machine Vision and Signal Analysis, University of Oulu, Oulu, Finland and School of Electronics and Information, Northwestern Polytechnical University, Xi'an, China. (e-mail: xiaoting.wu@oulu.fi, wuxt14@mail.nwpu.edu.cn)}
\thanks{E. Granger is with the Laboratoire d'imagerie, de vision et d'intelligence artificielle (LIVIA), Dept. of Systems Engineering, É\'Ecole de technologie sup\'erieure, Montreal, Canada. (e-mail: eric.granger@etsmtl.ca)}
\thanks{X. Feng is with the School of Electronics and Information, Northwestern Polytechnical University, Xi'an, China. (e-mail: fengxiao@nwpu.edu.cn)}
}
 

\ifCLASSOPTIONpeerreview
\begin{center} \bfseries EDICS Category: 3-BBND \end{center}
\fi
\IEEEpeerreviewmaketitle

\maketitle

\begin{abstract}
Visual kinship verification entails confirming whether or not two individuals in a given pair of images or videos share a hypothesized kin relation. As a generalized face verification task, visual kinship verification is particularly difficult with low-quality \emph{found} Internet data. Due to uncontrolled variations in background, pose, facial expression, blur, illumination and occlusion, state-of-the-art methods fail to provide high level of recognition accuracy. As with many other visual recognition tasks, kinship verification may benefit from combining visual and audio signals. However, voice-based kinship verification has received very little prior attention. We hypothesize that the human voice contains kin-related cues that are complementary to visual cues. 
In this paper we address, for the first time, the use of audio-visual information from face and voice modalities to perform kinship verification. We first propose a new multi-modal kinship dataset, called \emph{TALking KINship} (TALKIN), that contains several pairs of Internet-quality video sequences. Using TALKIN, we then study the utility of various kinship verification methods including traditional local feature based methods (e.g. LBP, LPQ, etc.) and statistical methods (e.g., GMM-UBM and i-vector), and more recent deep learning approaches (e.g., VGG, LSTM and ResNet 50). Then, early and late fusion methods are evaluated on the TALKIN dataset for the study of kinship verification with both face and voice modalities. Finally, we propose a deep Siamese fusion network with contrastive loss for multi-modal fusion of kinship relations. 
Extensive experiments on the TALKIN dataset indicate that by combining face and voice modalities, the proposed Siamese network can provide a significantly higher level of accuracy compared to baseline uni-modal and multi-modal fusion techniques. Our experiments show that we can obtain an EER of 40.1\% by using the voice modality, an EER of 32.5\% by using the face modality, and, by combining face and voice modalities with the deep Siamese fusion network, we can achieve an EER of 29.8\%. Experimental results also indicate that audio (vocal) information is complementary (to facial information) and useful for kinship verification.
 
\end{abstract}


\section{Introduction}



\IEEEPARstart{H}{uman} faces have abundant attributes that can implicitly indicate the family heredity from visual appearance. This phenomenon has been studied in psychology, with the aim of discovering how humans visually identify kin related cues from face~\cite{debruine2009kin, dal2010lateralization, dal2006kinin, wu2013male}. Driven by this research, the computer vision and machine learning communities have addressed this new problem --- \emph{kinship verification} from facial images. In this context, specialized techniques have been proposed to automatically verify whether two facial images have kin relation. This topic has received much attention since the first study on kinship verification from facial images by Fang~\emph{et. al}~\cite{fang2010towards} in 2010. 

In the literature, kinship verification is mainly explored from a perspective of visual appearance~\cite{fang2010towards}\cite{lu2014neighborhood}\cite{kohli2018supervised}, with most techniques based on still images. Just as people with kin relations tend to share common facial attributes, they may also share common voice attributes. In the genetic study domain, to determine how the human voice is passed down through generations, and to study the key factors influencing our voice, researchers from University of Nottingham carried out a pilot study on the heritability of human voice parameters\footnote{\url{https://www.nottingham.ac.uk/news/pressreleases/2016/january/help-the-scientists-find-out-why-you-sound-like-your-parents.aspx}}. Inspired by this study, we address, for the first time, the use of vocal information for kinship verification. Despite the long history of speech research, assessing kinship relation from voice has received very little attention in literature --- some studies have addressed potential performance degradation of \emph{automatic speaker verification} (ASV) when tested with the voice of persons with close kinship relation, such as identical twins~\cite{ariyaeeinia_test_2008, kunzel_automatic_2011}. 
Furthermore, many related applications, like expression recognition in affective computing, have benefited from using techniques that combine face and voice modalities~\cite{tzirakis2017end, chowdhury2018msu, liu2018towards, neverova2016moddrop}. In this paper, we hypothesize that fusing face and voice modalities captured in video sequences can improve the accuracy and robustness of systems for kinship verification. 

Audio-visual kinship verification has many potential applications ranging from social media analytics, forensics, 
surveillance and security to kin-related authentication. For instance, social media applications involve an overwhelming amount of data, including face images and videos. Automatic kinship verification could be used in semi-automatic organization of kin relations within different social relations, such as friends or colleagues. Another application of audio-visual kinship verification is to find missing children after some years, even when their appearance changes due to ageing, rather than expensive and invasive DNA test. It can also be employed in surveillance and security control in abnormal behavior detection. Through the analysis of surveillance footage video and verifying kinship relations, crime such as children kidnapping can be detected and kinship verification could be a decision support tool for forensic investigation. 
Automatic kinship verification system can also be used in kin related authentication. For instance, the United States Department allows people with relatives resided in the U.S. to enter as refugees~\cite{kohli2018supervised}. Audio-visual kinship verification can implement the real-time kin test with low cost. Audio-visual kinship analysis can also be used for automatic video organization and annotation.

This work focuses on kinship verification using audio-visual information. We investigate verification systems that allow for fusion of facial and vocal modalities to encode a discriminative kin information. The main contributions of this work are summarized as follows.
\begin{enumerate}
    \item Since no available kinship database is available for studying multi-modal kinship verification, we collected and analysed a new kinship database called \emph{TALking KINship} (TALKIN). It consists of both visual (facial) and audio (vocal) information of individuals captured talking in videos. We consider four kin relations: Father-Son (FS), Father-Daughter (FD), Mother-Son (MS) and Mother-Daughter (MD). 
    \item We consider sub-problems driven by the TALKIN database: kinship verification from facial images, voice, and from audio-visual information. We investigate the impact on performance (accuracy and complexity) when going from uni-modal to multi-modal cases. Benchmark results for uni-modal kinship verification are provided and then for several fusion methods are analysed and compared with state-of-the-art uni-modal and multi-modal fusion techniques.
    \item A deep Siamese fusion network with contrastive loss is proposed for audio-visual information fusion, to enhance the reliability of kinship predictions. Experiments show that the proposed fusion methods outperform baseline uni-modal and multi-modal fusion methods.
\end{enumerate}
This paper extends our preliminary investigation on audio-visual kinship verification~\cite{Talkin2019} in several ways. In particular: (1) a comprehensive analysis of related literature from the perspective of kinship verification, automatic speaker verification from close kin relations and multi-modal methods, for a more self-contained presentation; (2) a detailed description of the proposed and baseline methods for kinship verification based on face, voice and multiple modalities;  and (3) more proof-of concept experimental results and interpretations, including a detailed analysis of performance for audio vs. video based kinship verification.

The rest of this paper is organized as follows. Section~\ref{sec:related work} provides background and previous research related to existing kinship databases, proposed kinship verification techniques, automatic speaker verification for identical twins and multi-modal fusion applications. Section~\ref{sec:db} introduces the \emph{TALking Kinship} (TALKIN) dataset. In Section~\ref{sec:methodology}, kinship verification problem are presented from from a perspective of one modality (face \emph{vs.} voice) and multiple modalities (face \& voice). Techniques for both uni-modal and multi-modal kinship verification are presented. Section~\ref{sec:experiment setup}, describes the experiment methodology employed for performance evaluation. Finally, in  Section~\ref{sec:experiment} the experimental results are presented and discussed. 

\section{Related Work}
\label{sec:related work}

\subsection{Databases}

Since 2010, several kinship databases have been published: Cornell KinFace~\cite{fang2010towards}, UB KinFace~\cite{xia2011kinship,shao2011genealogical,xia2012understanding}, KinFaceW~\cite{lu2014neighborhood}, UvA-NEMO Smile~\cite{dibekliouglu2012you,dibeklioglu2013like}, TSKinFace~\cite{qin2015tri}, KFVW \cite{yan2018video} and FIW~\cite{robinson2018visual}. All these datasets address the general problem of kinship verification modeling using facial images or videos, but differ both in their exact task settings as well as the quality and quantity of data. We provide below a brief review of each dataset.

\textbf{Cornell KinFace}~\cite{fang2010towards} is the first kinship database that aims to verify kin relations using computer vision and machine learning methods. It includes 150 parent-child face image pairs of celebrities. The facial images were collected from the Internet by researchers at Cornell University, and represent therefore uncontrolled, \emph{in the wild} style data with no control over environments, cameras or poses. 

\textbf{UB KinFace}~\cite{xia2011kinship,shao2011genealogical,xia2012understanding} is the only kinship verification database that includes images of parents when they were both young and old. UB KinFace consists of two parts focused on Asian and non-Asian subjects, respectively. Each part has 100 groups of facial images. Each group has one image of child and one image for both young and old parent. Thus, in total there are 600 ($2\times 100 \times 3$) facial images. The resolution of images is 89 $\times$ 96.

\textbf{KinFaceW}~\cite{lu2014neighborhood} has two subsets, KinFaceW-I and KinFace-II. Images of each parent-child pair from KinFaceW-I are collected from different photos while image pairs in KinFaceW-II are from the same family photograph. The facial images are aligned according to eye position and cropped into size of 64 $\times$ 64.

\textbf{TSKinFace}~\cite{qin2015tri} database addressed the problem that children may partially seem like one parent and also partially look like the other parent. It consists 1015 tri-subject groups (Father-Mother-Child) totally.

\textbf{Families in the Wild} (FIW)~\cite{robinson2018visual} is the largest and most comprehensive visual kinship database with over 13,000 family photos of 1,000 families (with average of 13 of each family). Facial images are resized into 224 $\times$ 224.

\textbf{UvA-NEMO Smile}~\cite{dibekliouglu2012you,dibeklioglu2013like} database addresses the problem of video based kinship verification. It is collected under constrained environment with limited real-world variation that people make a smile face spontaneously and deliberately. To carry out the study of video based kinship verification from more complex environment, Yan~\emph{et al.} collected \textbf{Kinship Face Videos in the Wild} (KFVW)~\cite{yan2018video} database under unconstrained environment. It was collected from the TV show on the Internet with 418 pairs of facial videos. 

While the above databases cover multiple aspects of kinship verification from faces, no publicly available audio-visual kinship verification database exists. To explore the problem of kinship verification from face and voice modalities, we collected and analysed the TALKIN dataset (see Section~\ref{sec:db}).


\subsection{Kinship verification from faces}

Kinship verification from facial images was first addressed by Fang~\emph{et al.}~\cite{fang2010towards} in 2010. Since then, many works have been proposed and several competitions have been organized~\cite{lu2014kinship,lu2015fg,Robinson2017RFIW,robinson2017recognizing}. We briefly review below related works on visual kinship verification from still images and videos.

\subsubsection{Image-based verification:}
Initial works focused on feature based methods. Fang~\emph{et al.}~\cite{fang2010towards} extracted 22 facial features and selected the 14 most discriminative ones for classification; distance between two images is calculated and fed into \emph{K-nearest Neighbor} (KNN) and \emph{Support Vector Machine} (SVM) back-ends to verify the kin/non-kin relation. Yan~\emph{et al.}~\cite{yan2014prototype} proposed \emph{prototype-based discriminative feature learning} (PDFL) method to learn a feature representation from the labeled face in the wild (LFW) dataset without kin labels. Wu~\emph{et al.}~\cite{wu2016usefulness} extracted color texture features to study the importance of color in kinship verification problem. Besides the works on feature representation, metric learning also showed good performance. Lu~\emph{et al.}~\cite{lu2014neighborhood} proposed \emph{neighborhood repulse metric learning} (NRML) method which aims to repulse the images without kin relation and minimize the distance between images with kin relation. Liu~\emph{et al.}~\cite{liu2017status}, in turn, proposed \emph{status-aware projection metric learning} (SPML) method to solve the asymmetric problem as parent and child are considered with different status that parent is usually older than child. Finally, \emph{deep learning} \cite{Goodfellow-et-al-2016} shows high performance in the field of computer vision, kinship verification being no exception~\cite{zhang12kinship,lu2017discriminative}. Zhang~\emph{et al.}~\cite{zhang12kinship} proposed an end-to-end \emph{convolutional neural network} (CNN) architecture for kinship verification that uses a pair of two RGB images as input, and a softmax layer to predict the kinship relation. Compared with other state-of-the-art methods, such as \emph{discriminative multimetric learning} (DMML)~\cite{yan2014discriminative}, it yielded 5.2$\%$ and 10.1$\%$ improvement on KinFaceW-I and II, respectively. To further demonstrate the discrimination of CNN, Lu~\emph{et al.}~\cite{lu2017discriminative} presented \emph{discriminative deep metric learning} (DDML) method to learn a non-linear distance metric. The back-propagation algorithm was used to train the model where the distance between the positive pairs was narrowed and distance between negative pairs was enlarged.

\subsubsection{Video-based verification}
Video based kinship verification attracted less attention compared to still image based kinship verification. However, facial expression dynamics can provide useful information for kinship verification. It has been shown that people from the same family display similar facial expressions, such as anger, joy, or sadness~\cite{peleg2006hereditary}. Video-based kinship verification problem was studied by Dibeklioglu~\emph{et al.}~\cite{dibeklioglu2013like}. The authors localized 17 facial landmarks and used temporal \emph{Completed Local Binary Pattern} (CLBP) descriptors to describe the expressions. Combined with the spatial facial features, temporal CLBP features are fed into SVM to classify kin or non-kin relation. Then, Boutellaa~\emph{et al.}~\cite{Boutellaa_ICB16} proposed to use both shallow spatio-temporal features and deep features to characterize a dynamic face, which got a further improvement. Unconstrained video based kinship verification was recently proposed by Yan~\emph{et al.}~\cite{yan2018video}. They collected a new kinship database with videos in the wild condition. Several state-of-the-art metric learning algorithms were evaluated on video based kinship verification problem. Yet, previous work is mainly performed from computer vision domain. There is no study that has focused on kinship verification combining face and voice cues.

\subsection{Speaker verification for identical twins}

As far as we know, there are neither no specifically focused databases nor kinship verification studies using voice. Some related work exists within reliability assessment of speaker recognition. Voice from two different persons with a close kinship relation might be confusable. One special case --- voice of identical twins --- was addressed almost five decades ago~\cite{rosenberg_listener_1973}, when it was found to confuse listeners in same/different speaker discrimination. 

More recent studies, involving mostly automatic systems, have also demonstrated that voice of identical twins can be confusable also for automatic systems. The authors of~\cite{ariyaeeinia_test_2008} studied \emph{automatic speaker verification} (ASV) performance using voice of identical twins collected at a twin research institute in the UK. There are totally 49 identical twin pairs (40 female and 9 male pairs) involved. A Gaussian mixture model - universal background model (GMM-UBM)~\cite{reynolds_speaker_2000} with 2048 Gaussians was trained. They reported 0.4\% equal error rate (EER) when tested with all speakers, which degraded to 5.2\% EER when tested with twin voice. The EER increased from 2.8\% (all) to 10.5\% EER (twins) with short utterance.

The author of~\cite{kunzel_automatic_2011} studied the performance of a commercial forensic automatic speaker recognition with identical twin data. The author compared graphically likelihood ratio distributions and reported EERs from various experiments. Under matched-text condition, the author reported 0\% and 0.5\% EERs for males and females, respectively, when unrelated speakers were used as non-targets; these errors increased, respectively, to 11\% (male) and 19.2\% (female) when twins were used as non-targets. This 19.2\% was increased up to 48\% with mismatched texts. In summary, the tested automatic system experienced performance degradation for both genders and much worse for females. 


Besides observing the performance change of automatic systems, a number of studies focus on acoustic differences of twins. For instance, \cite{zuo_formant_2015} studies formant dynamics of 8 Shanghainese-Mandarin bilingual identical twin pairs, focused on common diphthong /ua/ found in both languages. The authors discovered that although very similar, identical twins did have significant differences in their formant dynamics. The authors constructed a simple \emph{linear discriminant analysis} (LDA) classifier formed from the first three formants (F1 to F3) and reported speaker classification rates between ~80\% to 90\%.

Despite the use of small datasets, the above review does suggest that voice of identical twins are potentially confusable by some listeners and ASV systems. While detrimental for ASV, the news are positive from the perspective of kinship verification: it looks possible to devise a system or a method that is sensitive to kinship cues in the human voice, to be used for detecting how closely two speakers are related. While identical twins are a rare special case in the general population, an interesting open question is how accurately kinship relations could be determined from voice for more common kinship relationships addressed in the visual kinship studies. One of the main aims to introduce our TALKIN database is to help answering this question.


\begin{table*}[htb!]
\renewcommand{\arraystretch}{1.3}
\caption{Main characteristics of existing datasets for kinship verification.}
\label{database}
\centering
\begin{tabular}{|c|c||c|c|c|c|c|}
\hline
\multicolumn{2}{|c||}{\bf{Database}} & \bf{Modalities} & \bf{Size} & \bf{Resolution ratio} & \bf{Family structure} & \bf{Controlled environment}\\
\hline
\hline
\multicolumn{2}{|c||}{Cornell KinFace~\cite{fang2010towards}} & Image & 150 pairs & $100\times 100$ & No & No\\
\hline
\multicolumn{2}{|c||}{UB KinFace~\cite{xia2012understanding}\cite{xia2011kinship}\cite{shao2011genealogical}} & Image & 200 groups & $89\times 96$ & No & No\\
\hline
\multirow{2}{*}{KinFaceW~\cite{lu2014neighborhood}} & KinFaceW-I & Image & 533 pairs & $64\times 64$ & No & No\\
\cline{2-7}
& KinFaceW-II & Image & 1000 pairs & $64 \times 64 $ & No & No\\
\hline
\multicolumn{2}{|c||}{TSKinFace~\cite{qin2015tri}} & Image & 1015 tri-subjects & $64 \times 64$ & No & No\\
\hline
\multicolumn{2}{|c||}{UvA-NEMO Smile~\cite{dibekliouglu2012you}\cite{dibeklioglu2013like}} & Video & 1240 videos & $1920\times 1080$ & No & Yes\\
\hline
\multicolumn{2}{|c||}{FIW~\cite{robinson2018visual}} & Image & 1000 family trees & $224\times 224$ & Yes & No\\
\hline
\multicolumn{2}{|c||}{KFVW~\cite{yan2018video}} & Video & 418 pairs of videos & $900 \times 500$ & No & No\\
\hline
\hline
\multicolumn{2}{|c||}{TALKIN (ours)} & Video \& Audio & 400 pairs of videos & 1920 $\times$ 1080 & No & No\\
\hline
\end{tabular}
\end{table*}

\subsection{Multi-modal methods}
Multi-modal fusion methods have successfully improved the recognition accuracy in many applications found in affective computing~\cite{tzirakis2017end}, person recognition~\cite{chowdhury2018msu}, large-scale video classification~\cite{liu2018towards} and gesture recognition~\cite{neverova2016moddrop}, because they can exploit complementary sources of information. Different sources of information are typically integrated through early fusion (feature level) or through late fusion (score or decision levels)~\cite{baltruvsaitis2018multimodal}. Feature-level fusion using concatenation or aggregation (\emph{e.g.}, \emph{canonical correlation analysis} or CCA~\cite{correa2010canonical}) is often considered to provide a high level of accuracy, although feature patterns may also be incompatible and increase system complexity. Techniques for score-level fusion using deterministic (\emph{e.g.}, average fusion) or learned functions are commonly employed, but are sensible to the impact of score normalization methods on the overall decision boundaries 
and the availability of representative training samples. Despite reducing the information content about modalities, techniques for decision-level fusion (e.g., majority voting) can provide a simple framework for combination, although limitations are placed on decision boundaries due to the restricted operations that can be performed on binary decisions.

In the deep learning literature, Neverova~\emph{et al.}~\cite{neverova2016moddrop} proposed a multi-scale and multi-modal early fusion method --- \emph{multimodal dropout} (ModDrop) --- for gesture recognition problems. First, the weights of each modality are pre-trained. Then, a gradual fusion method is proposed by randomly dropping separate channels to learn cross-modal correlations while preserving uni-modality specific representation.
Liu~\emph{et al.}~\cite{liu2018towards}, in turn, introduced \emph{multi-modal factorized bi-linear pooling} (MFB)~\cite{yu2017multi} method to combine visual and audio representations for video-based classification. In affective computing applications, Tzirakis~\emph{et al.}~\cite{tzirakis2017end} proposed an end-to-end multimodal deep NN for emotion recognition. Visual and speech modalities are first trained separately to speed up the fusion training phase. Then, the fusion network is trained in an end-to-end fashion. 
Concerning late fusion, authors of ~\cite{chowdhury2018msu} considered multiple score fusion techniques for indoor surveillance person recognition. Experimental results showed the efficiency of multimodal methods over the unimodal approaches.

To sum up, prior results in literature suggest that improvements in accuracy and robustness can be obtained by using multi-modal methods over uni-modal techniques. To improve the accuracy and robustness of kinship verification, we therefore investigate algorithms for the fusion of face and voice modalities. 
To the best of our knowledge, our work is the first attempt to study the kinship verification from both visual and audio information.


\section{The TALKIN Database}
\label{sec:db}
\begin{figure*}[!ht]
\centering
\includegraphics[width=7in]{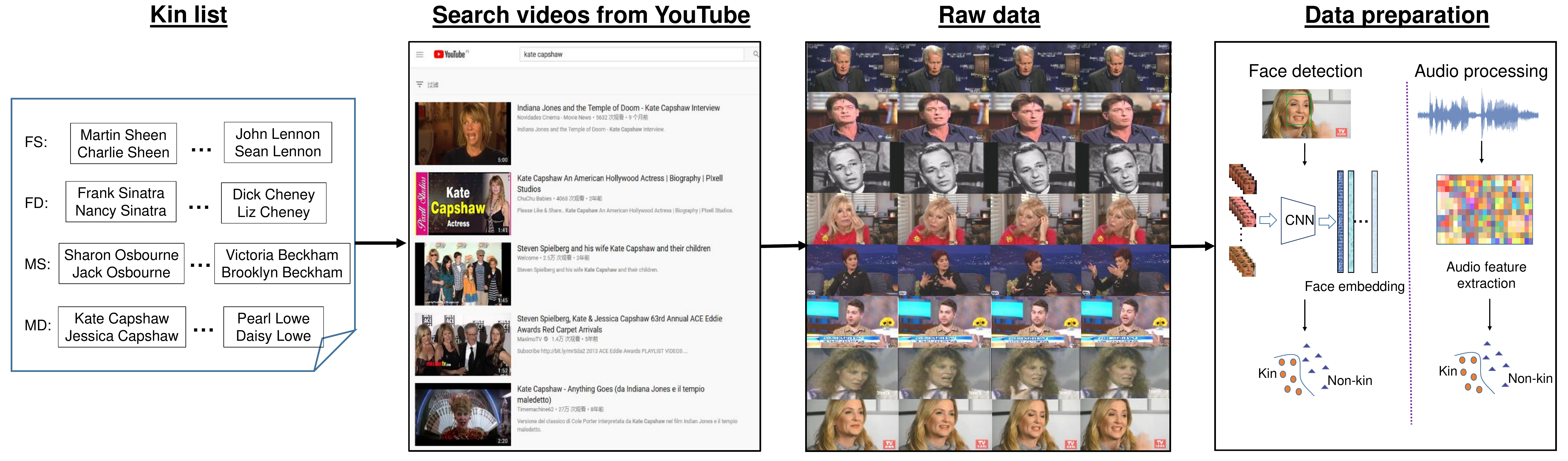}
\caption{The pipeline employed to collect and analyse the TALKIN database. \textsl{Kin list.} A list of candidates is first summarized as the preparatory work. \textsl{Search videos from YouTube.} The facial videos with vocal information are searched and downloaded from YouTube. \textsl{Raw data.} Several samples from the TALKIN are shown. From the top to the bottom, there are father-son, father-daughter, mother-son and mother-daughter. As can be seen, TALKIN database has varieties of environment, background, pose and illumination. \textsl{Data preparation.} Videos and audio are pre-processed. Then the embeddings for video and audio are extracted for the task of kinship analysis.}
\label{Talkin pipeline}
\end{figure*}

In this section, we describe our new kinship database called \emph{TALking KINship} (TALKIN). Compared with existing kinship databases with facial videos (UvA-NEMO Smile~\cite{dibekliouglu2012you}\cite{dibeklioglu2013like} and KFVW~\cite{yan2018video}), TALKIN contains both videos and audio under unconstrained environment. It contains several videos of subjects talking in the wild environment (under unconstrained background, illumination and recording condition). The purpose of collecting our new database is to investigate the problem of audio-visual kinship verification in the wild. A comparison of TALKIN with existing kinship databases is shown in Table~\ref{database}.

\subsection{Data collection pipeline}

The overall collection pipeline of the TALKIN dataset is shown in Fig.~\ref{Talkin pipeline}. 

\textbf{Step 1. List of celebrities or family TV shows.} 
First, we prepared a name list that we intend to obtain videos from. The target amount of video for each relation is 100 pairs of clips. Most of the list is formed by celebrities, such as musicians, actors, politician~\emph{et al.}, with rest of it from TV series involving family interactivity (non-celebrities).

\textbf{Step 2. Downloading videos from YouTube.} 
We downloaded the videos from YouTube by searching the name of celebrities or TV series. To avoid biases encountered in some of the previous kinship databases~\cite{miguel2016,dawson2018same}, we collected parent's videos and child's videos from \emph{different} video clips corresponding to different backgrounds and recording conditions.

\textbf{Step 3. Data preparation.} After we getting the raw data from the web, we did data pre-processing. For face detection and alignment, we employed \emph{Multi-task Cascaded Convolutional Networks} (MTCNN) algorithm~\cite{MTCNN} to detect 5 face landmarks in every frame of the video. Finally, the videos are cropped and aligned according to the landmarks. The facial regions are then re-sized into 224$\times$224. Both hand-crafted features and deep features are extracted to represent each individual. To represent the audio information, we directly extracted audio from the video clips. The sample rates are all set to 44.1 kHz. Three standard techniques in the speech field, namely GMM-UBM, i-vectors and Deep Neural Network, are used for text-independent kinship analysis.

\begin{table*}[ht]
\caption{The ethnicity distribution (\%) of TALKIN dataset.}
\label{table: ethnicity}
\centering
\begin{tabular}{|c|c|c|c|c|c|c|c|c|}
\hline
British & American & French & Australian & Chinese & Dutch & Italian & Swedish & Turkish \\ \hline
56.50  & 33.50   &  6.50 &   2.00      &  0.50  & 0.25 & 0.25 &  0.25 &  0.25 \\ \hline
\end{tabular}
\end{table*}

\subsection{Parameters of the dataset}

The TALKIN dataset focuses on four kin relations: Father-Son (FS), Father-Daughter (FD), Mother-Son (MS) and Mother-Daughter (MD), with 100 pairs of videos (with audio) for each relation. As all the data originates from uncontrolled Internet sources, the speech contents vary from subject to subject and video to video, making the voice-related sub-task \emph{text-independent kinship verification}, analogous with text-independent speaker verification. That is, the task is to verify kinship relations regardless of what was said between individuals.

TALKIN incorporates a wide range of backgrounds, recording environments, poses, occlusions and ethnicities. Table~\ref{table: ethnicity} shows the distribution of ethnicity in TALKIN. The distribution is count by kin pair rather than individuals, in case that one parent might appear multiple times with more than one kid. Note, however, that we exclude mixed-race trials, \emph{i.e.} the parent and child in a trial has the same ethnicity. The dataset has two parts: video and audio. The length of the video varies from 4.032 seconds to 15 seconds with a resolution of $1920 \times 1080$. Audio is extracted from video files. Besides the varied text content, the audio files contain substantial channel variations (\emph{e.g.} due to differing recording devices). Some of them also contain reverberation and additive noise.

\section{Methodology}
\label{sec:methodology}

\begin{figure}[htb!]
\centering
\subfigure[]{\includegraphics[width=3.5in]{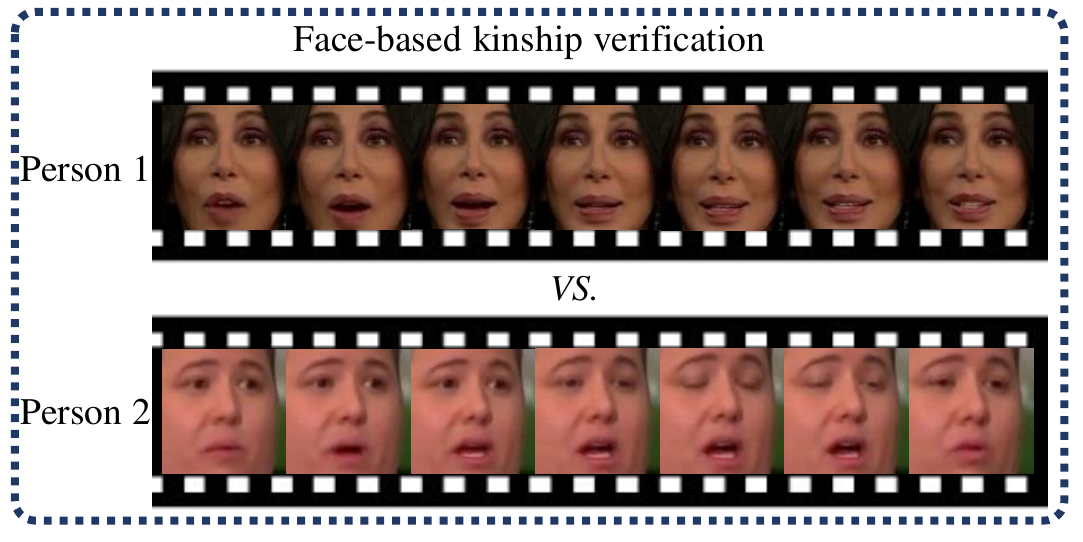}
\label{subfig:face-modal structure}}
\hfil
\subfigure[]{\includegraphics[width=3.5in]{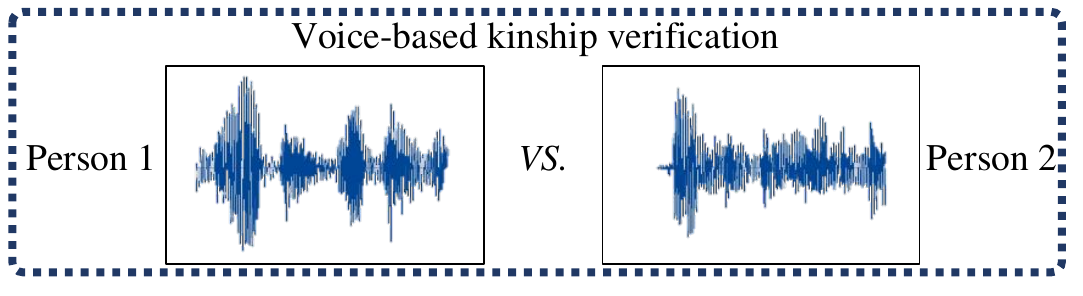}
\label{subfig:voice-modal structure}}
\hfil
\caption{Kinship verification from a single modality. In~\ref{subfig:face-modal structure} we determine whether two persons have kin relation from facial videos, while  in~\ref{subfig:voice-modal structure} from voice.}
\label{fig: uni-structure}
\end{figure}

In this section, we will focus on the kinship verification problem using both uni-modal (face \emph{vs.} voice) and multi-modal methods. Fig.~\ref{fig: uni-structure} shows examples of signal employed for kinship verification from face and voice modalities. Fig.~\ref{fig: multi-structure} illustrates the architectures proposed for uni-modal systems. We address kinship detection as a \emph{hypothesis testing} problem -- given a pair of signals (a pair of video sequences or speech utterances), say $(S_1,S_2)$, the task is to evaluate support for two mutually exclusive hypotheses, \emph{null} hypothesis $H_0$ and \emph{alternative} hypothesis, 
\begin{equation}
\begin{cases}
    H_0: & \text{$S_1$ and $S_2$ are of the same kin}\nonumber  \\
    H_1: & \text{$S_1$ and $S_2$ are of different kin.}\nonumber
\end{cases}
\end{equation}
In practice, we represent $S_1$ and $S_2$ using frame-level feature vectors that are then used to derive recording-level representations of fixed size (regardless of the number of frames). Kinship score -- a numerical indicator with higher values associated with stronger support in favor of $H_0$ -- is then obtained by computing similarity score between the feature representations. We consider both hand-crafted and data-driven (learned) feature representations and similarity scoring techniques. The  following three subsections present methods for face-based, voice-based feature representations and data fusion, respectively.

\begin{figure*}[htb!]
\centering
\includegraphics[width=6in]{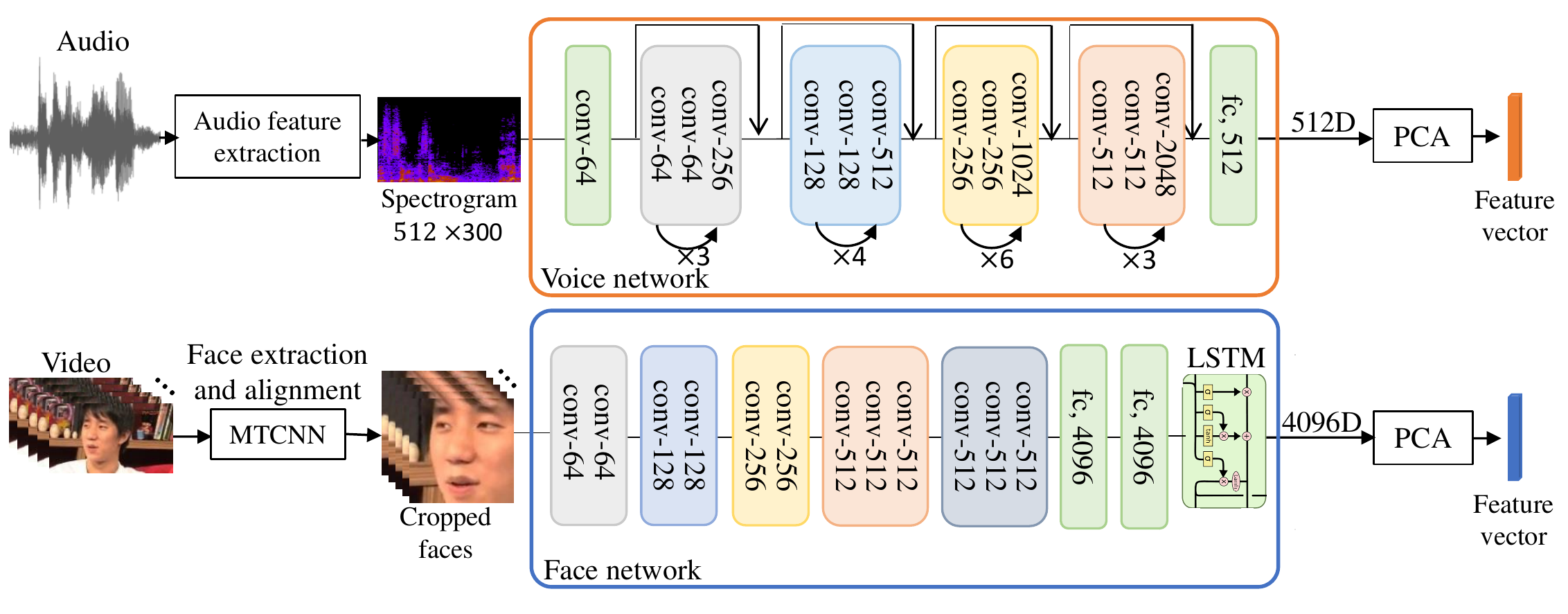}
\caption{Architecture of the proposed uni-modal methods. Both the face and voice modalities use the similar but specialized convolutional architectures trained for each modality separately. The convolutional layers learn data-driven feature extractors. Face network is formulated by VGG embeddings with a layer of LSTM to integrate the spatial-temporal information, while LSTM is trained with back propagation with contrastive loss. The voice network is based on \emph{ResNet-50} that is pre-trained with \emph{VoxCeleb} dataset. We optimize the last layer of \emph{ResNet-50} with the Siamese architecture. This way, we obtain 512- and 2880-dimensional discriminative voice and face embeddings, respectively. Their dimensionalities are further reduced by principal component analysis (PCA). After we get the reduced dimensional feature, distance metric is calculated to classify whether they have kin relation or not.
}
\label{fig: uni-architecture}
\end{figure*}

\subsection{Face-based kinship verification}
\label{subsec:face-based kinship}

We employed both hand-crafted and deep feature representations for facial kinship verification. In particular, we adopted image-based representations produced with \emph{binarized statistical image feature} (BSIF)~\cite{bsif_icpr12}, \emph{local phase quantization} (LPQ)~\cite{lpq_isp08}, and \emph{local binary pattern} (LBP)~\cite{ojala1996comparative,ahonen2006face} descriptors. These features are extracted from each video frame and averaged over the frames to represent a video sequence. In addition, we considered \emph{local binary patterns from three orthogonal planes} (LBP-TOP)~\cite{zhao2007dynamic} that is more naturally suited for representation of faces over multiple frames in a video. Besides these conventional hand-crafted features, we further included state-of-the-art deep Siamese architecture. The Siamese network is a pair-wise match, where feature representations are extracted  through metric learning. To assess the kinship similarity score between two video sequences, we computed \emph{cosine similarity} measure between two feature representation vectors, $\Vec{x_1}$ and $\Vec{x_2}$: 
\begin{equation}
\label{eq:cosine}
\text{sim}(\Vec{x_1},\Vec{x_2})=\frac{\Vec{x_1} \cdot \Vec{x_2}}{\|\Vec{x_1}\|\cdot\|\Vec{x_2}\|} .
\end{equation}
A threshold is applied to $\text{sim}(\Vec{x_1},\Vec{x_2})$ to determine whether two inputs have a kin relation.

\subsubsection{Image-based representation}
In~\cite{wu2016usefulness}, the authors demonstrated the effectiveness of HSV color space for kinship verification problem. We first converted the facial images into HSV color space. We considered several descriptors for extracting the features from the facial images. BSIF is a binary texture descriptor that uses a small set of natural images~\cite{hyvarinen2009natural} as a training set to learn filters. LPQ is a blur-invariant image texture descriptor. LBP shows its effectiveness in face analysis. It computes a binary code for each pixel in an image. The binary patterns are counted into a histogram to represent the image texture. 
\subsubsection{Video-based representation}
LBP-TOP is an extension of LBP, in which the local binary patterns are extracted from three orthogonal planes of a frame sequence: XY, XT and YT, where X and Y denote the spatial coordinates and T means the time coordinate. For a video or a sequence of image, it can be viewed as a stack of XY planes in axis T, XT planes in axis Y and YT planes in axis X. LBP-TOP extracts features from each separate plane and concatenates them into one feature vector.
\subsubsection{Face network} While the above hand-crafted face descriptors have the benefits of being simple and interpretable, they are not specifically optimized for kinship cue representation. Similar to other visual pattern classification tasks, we expect substantially better results by leveraging from data-driven approaches that are directly optimized for a given task. To this end,  we implemented the VGG-Face~\cite{Parkhi15} CNN cascaded with an Long Short-Term Memory (LSTM)~\cite{hochreiter1997long} network for the facial representations. VGG-Face network is trained on a large face dataset with 2.6 million images of over 2662 people~\footnote{\url{http://www.robots.ox.ac.uk/~vgg/software/vgg\_face/}}. 
This network has shown interesting performance on face verification using both images and videos. Furthermore, it also shown the effectiveness of kinship verification with constrained facial videos~\cite{Boutellaa_ICB16}. As shown at the top of Fig.~\ref{fig: uni-architecture}, it consists of 13 convolution layers, each followed by \emph{rectified linear unit} (ReLU). Some of them are also followed by max pooling operator. The last two layers are FC layers that have 4096 outputs. 
We fed the facial frames one by one and collected the deep features from layer fc7~\cite{Boutellaa_ICB16}. 
A layer of LSTM with 4096 cells is stack on the basis of VGG-Face descriptor and trained to integrate the spacial information to spatial-temporal features. The network is trained in `Siamese' fashion using \emph{contrastive loss}~\cite{li2016kinship}. Here, the contrastive loss is defined as:
\begin{equation}\label{contrastive}
\mathcal{L}=\frac{1}{2N}\sum_{n=1}^{N} (y_n d_n^{2} + (1-y_n)\max(M-d_n,0)^2),
\end{equation}
where threshold $M$ denotes \emph{margin}, $N$ is the mini-batch size, $d_n=\left \| \Vec{a_n}-\Vec{b_n} \right \|^{2}$, $\Vec{a_n}$ and $\Vec{b_n}$ denote two sample feature vectors that are collected from the last state of LSTM, $y_n$ is the label of the sample pair. $y_n$ equals 1 when the inputs have the kin relation and $y_n$ equals 0 the otherwise. 



\subsection{Voice-based kinship verification}
\label{sub-sec:voice-based kinship}

As for the voice modality, we adopted three methods from the related task of \emph{automatic speaker verification} (ASV). Two of them, \emph{Gaussian mixture model --- universal background model} (GMM-UBM) \cite{reynolds_speaker_2000} and \emph{identity vector} (i-vector) \cite{dehak2011front}, are standard statistical classifiers while the last one uses deep learning.

\subsubsection{GMM-UBM} 
We first trained UBM from a training set of disjoint speakers to those used for kinship scoring. The UBM, denoted here by $\Vec{\theta}_\text{ubm}$, models speaker-independent distribution of the MFCC features. It serves both as \emph{a prior} model to obtain speaker-dependent models via \emph{maximum a posterior} (MAP) adaptation, and as the likelihood model for the alternative hypothesis modeling. If we denote the MFCC sequence of a test utterance by $X=\{\vec{x}_1,\vec{x}_2,\dots,\vec{x}_T\}$ and the speaker model of $i$th speaker by $\vec{\theta}_i$, the detection score is given by the log-likelihood ratio (LLR) $\ell=\log p(X|\vec{\theta}_i) - \log p(X|\vec{\theta}_\text{ubm}$). When the speaker identities of $X$ and $\vec{\theta}_i$ are the same, LLR score is high (relative to situation when their identities differ).

For kinship modeling, speakers with a positive kin relation share the same source identity (kin label). When we evaluate a particular kin hypothesis, we compare the MFCCs of the test speaker against the speaker model of \emph{another} speaker. A positive kinship trial occurs when the test speaker (source of $X$) and the reference speaker (source of $\vec{\theta}_i$) have a positive kin relation (\emph{e.g.} mother-son). Pairs with no kin relation constitute negative trials. From this perspective, GMM-UBM is used exactly the same way as in ASV, though the trial labels are defined differently. Note that in kinship verification, all the compared speaker pairs have disjoint identities.


\subsubsection{I-vector based method} I-vector  \cite{dehak2011front,Kenny12-small-footprint} is a compact representation of a speech recording. It is extensively used in speaker and language recognition to represent speech utterances of different lengths as fixed-dimensional embeddings. Akin to GMM-UBM, the i-vector paradigm builds upon GMM modeling of short-term spectral observations. Unlike GMM-UBM, however, the i-vector model leverages from  statistical redundancy across different recordings by imposing subspace constraints to the mean vectors of a GMM. In specific, the model assumes that the mean vector of the $c$th Gaussian in recording $r$, denoted by $\vec{\mu}_{c,r}$, 
can be expressed as,
\begin{equation}
\label{eq:supervector}
\vec{\mu}_{c,r} = \vec{m}_c + \vec{T}_c\vec{\omega}_r,
\end{equation}
where 
$\vec{m}_c$ is recording-independent mean (from the UBM), $\vec{T}_c$ is recording-independent factor loading matrix and 
%
$\vec{\omega}_r$ is a latent random variable with a normal standard prior, $\vec{\omega}_r \sim \mathcal{N}(\vec{0},\vec{I})$. Here, $(\vec{m}_c, \vec{T}_c)_{c=1}^C$, where $C$ is the number of Gaussians, are the model hyper-parameters trained from offline data (\emph{i.e.} speakers disjoint from those used in kinship training/testing). The i-vector itself, denoted by $\vec{w}_r$, is the \emph{posterior mean} of $\vec{\omega}_r$ conditioned on recording-specific \emph{Baum-Welch sufficient statistics} collected using the UBM. We point the interested reader to \cite{Kenny12-small-footprint} for further details.

A key point is that an i-vector serves as a recording-level feature vector to compactly represent stationary recording-level cues embedded in the GMM means. Importantly, the i-vector extractor is trained in an \emph{unsupervised} way: the training of $\{\vec{m}_c\}$ (the UBM means) and $\{\vec{T}_c\}$ (the factor loading matrices) are done via dedicated \emph{expectation-maximization} (EM) approach that requires no training labels. This makes the i-vector itself agnostic to a given classification task at hand. To be useful for a given task (here, kinship verification), one further trains a back-end classifier with labeled i-vectors (here, with known family identity). The support towards positive kinship hypothesis for a pair of i-vectors (\emph{e.g.} hypothesized mother-son) can be then evaluated with the back-end classifier. After a number of tentative experiments, we ended up to \emph{linear discriminant analysis} (LDA) trained with family labels, followed by cosine scoring.


\begin{table*}[htb]
\centering
\caption{Summary and comparison of uni-modal methods on TALKIN dataset.}
\label{Table: methods summarize}
\begin{tabular}{|l|l|l|l|l|l|l|}
\hline
\multirow{2}{*}{Modality} & \multirow{2}{*}{Techniques} & \multirow{2}{*}{Operation} & \multirow{2}{*}{External data  usage} & \multicolumn{3}{c|}{Kinship verification procedure}
\\ \cline{5-7}
& & & & \begin{tabular}[l]{@{}c@{}}Representation extraction\\ (Frozen layers)\end{tabular} & Layers for fine-tune & Kinship classifier
\\ \hline
\multirow{5}{*}{Video} & BSIF & Average & - & - & - & \multirow{5}{*}{Cosine score}
\\ \cline{2-6}
& LPQ & Average & - & - & - &
\\ \cline{2-6}
& LBP & Average & - & - & - &
\\ \cline{2-6}
& LBP-TOP & \begin{tabular}[l]{@{}l@{}}From three\\orthogonal planes\end{tabular} & - & - & - &
\\ \cline{2-6}
& VGG+LSTM & Data-driven & VGGFace~\cite{Parkhi15} & VGG-Face & LSTM &
\\ \hline
\multirow{3}{*}{Audio} & I-vector & Data-driven & No-Within TALKIN\footnotemark[3] & \multicolumn{2}{l|}{Train UBM from scratch} & LDA+Cosine score
\\ \cline{2-7}
& GMM-UBM & Data-driven & No-Within TALKIN\footnotemark[3] & \multicolumn{2}{l|}{Train UBM and T matrix from scratch} & \begin{tabular}[l]{@{}c@{}}Log-likelihood\\ ratio (LLR)\end{tabular} 
\\ \cline{2-7} 
& ResNet-50 & Data-driven & Voxceleb2~\cite{Chung18b} & \begin{tabular}[l]{@{}l@{}}Layers except for last\\two layers\end{tabular} & Last two layers & Cosine score
\\ \hline
\end{tabular}
\end{table*}

\begin{figure*}[htb!]
\centering
\includegraphics[width=5.5in]{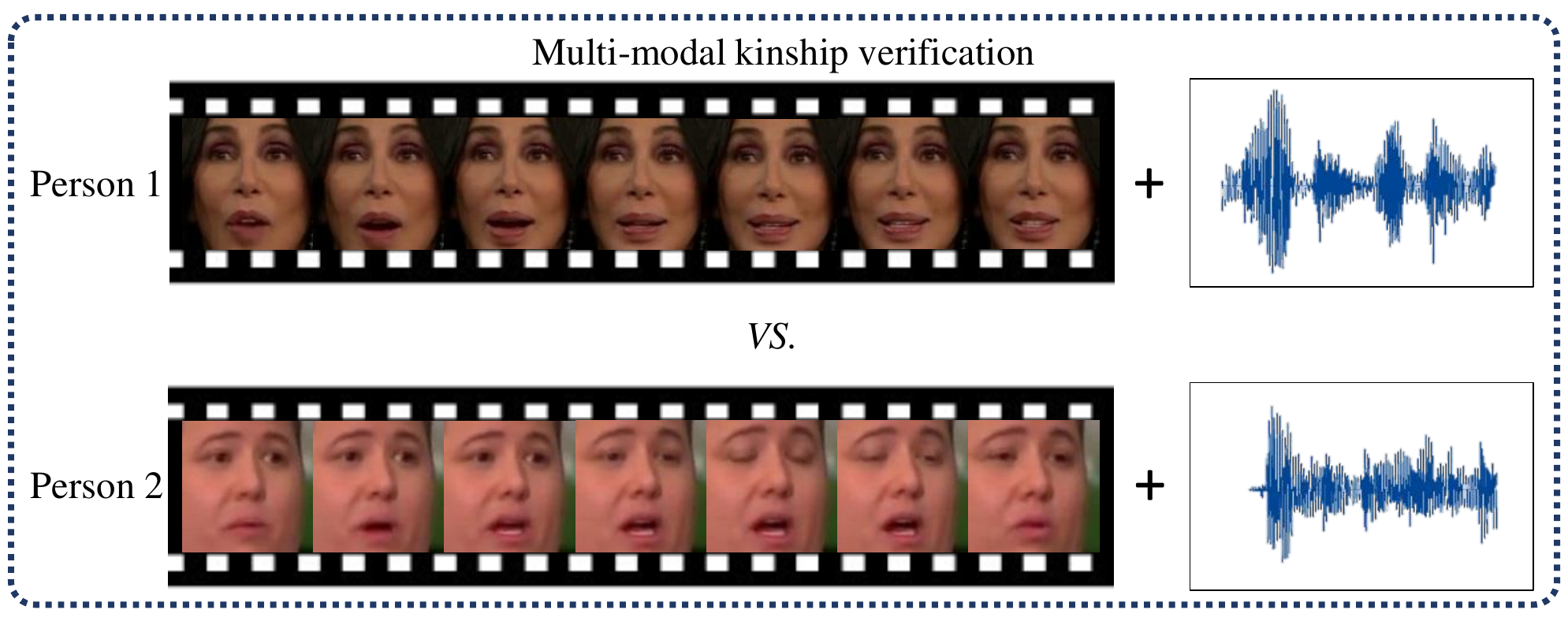}
\caption{Kinship verification from both face and voice modalities. We propose to fuse both visual information from face appearance and dynamics and vocal information to form a more complementary feature of one person.}
\label{fig: multi-structure}
\end{figure*}



\subsubsection{Voice network} Both the GMM-UBM and the i-vector methods are built upon fixed acoustic feature extractor (MFCC extractor), followed up by data-driven recording-level representation learning and back-end scoring. Even if both techniques have been successful in a number of speech-related tasks, one may question the usefulness of a fixed acoustic front-end. To this end, we wanted to further replace the i-vector embedding with a deep neural network model that uses convolutive models to extract features from the spectrogram, instead. In specific,
we rely on a pre-trained \emph{ResNet-50} model trained from a very large speaker verification dataset called VoxCeleb2~\cite{Chung18b}. We then fine-tune with TALKIN data to get feature embedding from it for audio based kinship verification, where we fix the convolutional layers and finetune the last fully connected layer.


The audio samples are first converted into single-channel and down sampled to 16 kHz to be consistent with VoxCeleb2. Then the audio samples are segmented into 3-second chunks. A Hamming-window of duration 25ms and 10ms step is applied on the audio. Following ~\cite{Chung18b}, spectrograms with the size of 512 frequency bins $\times$ 300 frames are extracted. After performing mean and variance normalization on the frequency bin of the spectrum, the normalized spectrograms are fed into the ResNet-50. Similar to the face network, to pull positive pairs (with kin relations) together and push negative pairs (without kin relations) away, the voice network is established as a Siamese network with contrastive loss at the end.

The overall uni-modal kinship verification methods are summarized in Table~\ref{Table: methods summarize}.

\footnotetext[3]{Disjoint speakers from those used in kinship training and scoring}


\subsection{Multi-modal kinship verification}
Up to this point, we have considered the visual and voice modalities in isolation from each other. In this sub-section we study effective ways to combine the modalities, where audio-visual kinship verification problem is illustrated in Fig.~\ref{fig: multi-structure}. This includes
introducing a novel deep Siamese network for the fusion of the two modalities, 
and the use of traditional early and late fusion strategies.
\subsubsection{Baseline fusion methods}
Two baseline methods for multi-modal kinship verification, early (feature) level and late (score) level fusion methods, are applied. For the early fusion method, after extracting features from face and voice network, 
PCA is used to make it consistent size for video and audio. Z-score normalization is used to normalize video and audio features separately. 
Then the video and audio features are concatenated together into one feature vector as the fused feature. Cosine similarity is calculated to classify whether they have kin relation. 


For the late fusion method, the evaluation for the video based and audio based kinship verification are performed separately, with corresponding match score $S_1$ and $S_2$. Then, the average score is selected as the fused score.
\subsubsection{A Siamese network for A-V fusion}

\begin{figure}[t!]
\centering
\includegraphics[width=3.5in]{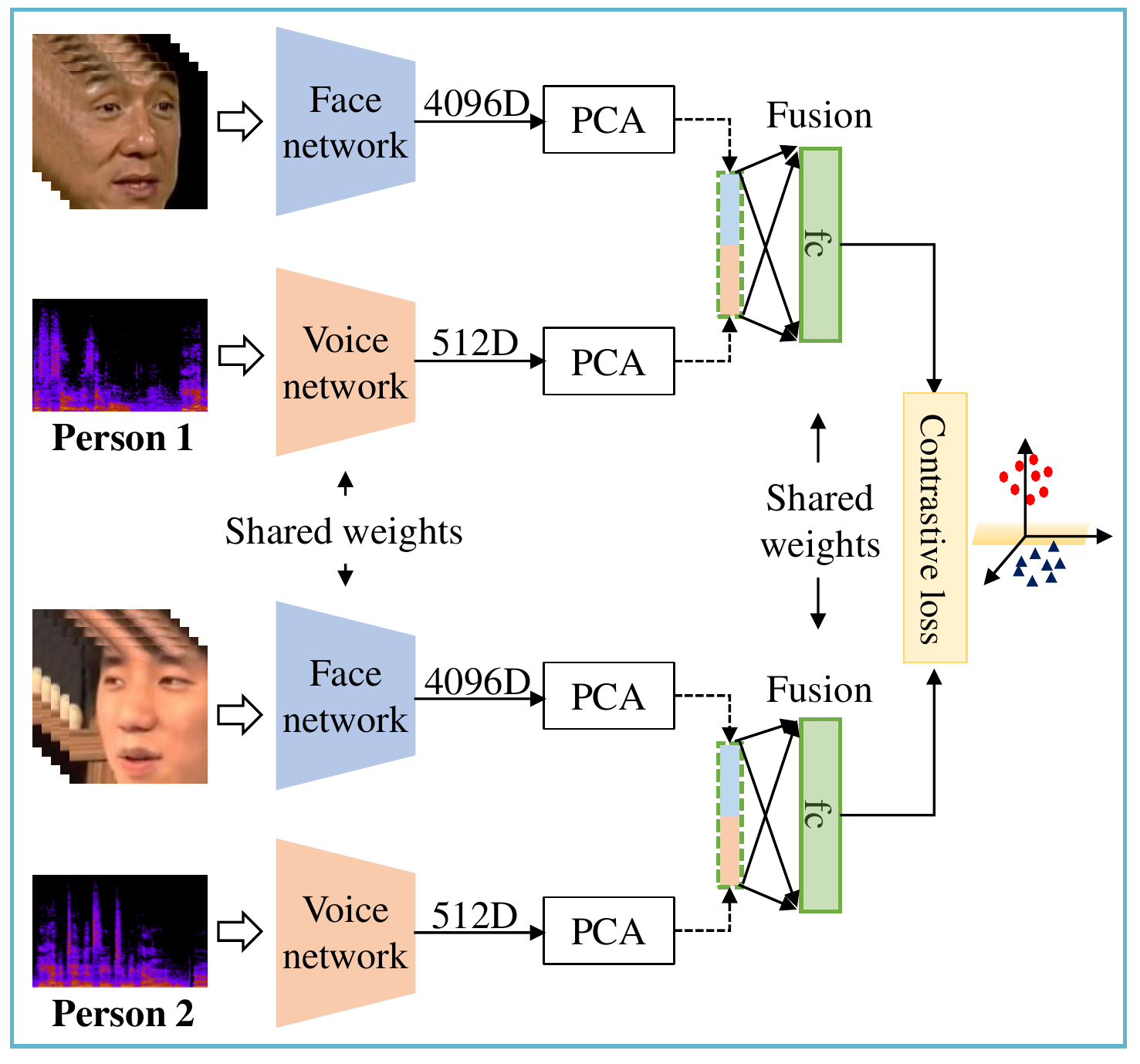}
\caption{Architecture of the proposed deep Siamese fusion network. The facial and vocal feature are extracted from the face and voice networks, respectively, that share the same parameters as Fig.~\ref{fig: uni-architecture}. After PCA, the concatenation of facial and vocal feature is connected with a fully connected layer to learn the fusion rule. It is trained in a Siamese fashion with pair-wise input with contrastive loss at last. Then the fully connected layer is represented as the fused feature for one subject.}
\label{fig: multi-architecture}
\end{figure}

The overall architecture of the deep Siamese network is shown in Fig.~\ref{fig: multi-architecture}. It is trained to evaluate pair-wise similarities based on face and voice modalities. In a particular implementation, we fine-tune the VGG-Face~\cite{Parkhi15} CNN cascaded with an LSTM network for the face modality, which is described in detail in subsection~\ref{subfig:face-modal structure}. For the voice modality (described in subsection~\ref{subfig:voice-modal structure}) extracted from videos, we fine-tune a ResNet-50 with TALKIN which is pre-trained on VoxCeleb2~\cite{Chung18b}. For each voice and face network, we use contrastive loss to learn the intra-class similarity and inter-class dissimilarity among subjects. 

After training the face and voice networks, we collected their features -- 4096 features 
from the face network and 512 features from the voice network. To make the dimensional balance of both facial and vocal representations, we applied PCA to reduce both facial and vocal feature dimensions into 130. Then they are concatenated into a 260-dimensional feature, followed by a FC layer with 260 nodes. During the training procedure, our system is trained on TALKIN, using backpropagation and  contrastive loss to learn the correlation between parent and child based on audio visual modalities, which has no family overlap between training and testing procedure. 
By adding contrastive loss during the fusion part, we can automatically learn the fusion rule for kinship verification to narrow the distance between pairs with a kin relation, and to enlarge the distance between the negative pairs. After training the network, the feature extracted from the added FC layer is viewed as fusion feature of one facial video and audio signal. Then, the cosine similarity $\text{sim}(\vec{x}_1,\vec{x}_2)$ is calculated to represent the distance between two inputs (\emph{e.g.} parent and child represented by feature vectors  $\vec{x}_1$ and $\vec{x}_2$). A threshold is applied to $\text{sim}$ to determine a kin relation.

\section{Experimental setup}
\label{sec:experiment setup}
The TALKIN dataset is used to evaluate the performance of uni-modal and multi-modal kinship verification. For each kin relation ---FS, FD, MS and MD--- there are 100 pairs of videos with a positive kin relation. Likewise, we randomly generate 100 pairs of videos without any kin relation as the negative pairs. Thus, for each sub-task, we have 100 pairs of positive and 100 pairs of negative pairs. We use 5-fold cross-validation setup in our experiment: for a given test fold of 40 pairs, we train a kinship detector from the held-out 160 pairs. There is no family overlap between the 5 folds.

\subsection{Parameter setup of methods}
\subsubsection{Hand-crafted features}
We employed the following image-based feature representations: BSIF, LPQ and LBP. We averaged these frame-by-frame features to represent each video by a single feature vector. The facial frames are first converted into HSV color space~\cite{wu2016usefulness} with size of 64 $\times$ 64 $\times$ 3. For BSIF feature extraction, images are divided into non-overlapping 32 $\times$ 32 blocks in each color channel. Each block is represented using 256 features and the whole face with 256 $\times$ 4 $\times$ 3 = 3072 features. For LPQ feature extraction, images are divided into non-overlapping 32 $\times$ 32 blocks in each color channel. Each block is represented using 256 features, leading to 3072-dimensional (256 $\times$ 4 $\times$ 3) feature representation for the whole face. For LBP feature extraction, the images are divided into non-overlapping 16 $\times$ 16 blocks in each color channel. The parameters of LBP are: the radius is set as 1 and the sampling number is 8. 59 histogram values are used to represent each block. Thus, each facial image is represented using 59 $\times$ 16 $\times$ 3 = 2832 features. Furthermore, we also evaluated the video representation, LBP-TOP. In the experiments, the frames are converted into gray scale. Then the face frames are divided into 56 $\times$ 56 non-overlapping blocks. All features extracted from each block volume are connected to represent the appearance and motion of the kinship video. The radius is 1. For each block volume, we extracted 59 histogram features in XY, XT and YT planes, respectively. Thus, one video can be represented as a 59 $\times$ 3 $\times$ 16 = 2832 face features. At last, we computed the cosine similarity between two facial features.
\subsubsection{Face network}
We fed the facial frames one by one with size of 90 $\times$ 224 $\times$ 224 $\times$ 3. The network is trained with 3 epochs with mini batch size of 40. Learning rate is set to $10^{-5}$. After collecting features from the last state of LSTM, PCA is performed to reduce the dimension into 110.
\subsubsection{GMM-UBM \& I-vector}
We used MSR Identity Toolkit~\cite{msr-identity-toolbox-v1-0-a-matlab-toolbox-for-speaker-recognition-research-2} to implement the GMM-UBM and i-vector methods. For both GMM-UBM and I-vector methods, we extract 12 Mel-frequency cepstral coefficients (MFCCs) from the audio samples with frame size of 256 and sample rate of 44.1 kHz. The UBM is trained with 128 Gaussian components. At last, we got i-vectors with dimensionality of 100. We used LDA to reduce the number of dimensions further down to 79 dimensions.
\subsubsection{Voice network}
The voice network is pre-trained on VoxCeleb2 dataset. Then we fine-tune the last two layers of network with learning rate of $10^{-3}$. The network is trained with mini batch size of 40 for 10 epochs. After training the network, audio features are extracted from the last fully connected layer of dimensionality of 512. PCA is performed to reduce the feature dimension into 144.
\subsubsection{Baseline fusion methods}
During both early fusion and late fusion, we kept the 144 dimensions of both video and audio features with PCA.
\subsubsection{Siamese network for A-V fusion} The fusion network is trained with mini batch size of 40 for 5 epochs. The learning rate is $10^{-5}$. Further, face network and Siamese network for A-V fusion are performed on TensorFlow~\cite{abadi2015tensorflow} with Nvidia Tesla P100 GPU running CentOS 7.6.1810, while voice network is performed on MatConvNet~\cite{vedaldi15matconvnet}.

\subsection{Performance evaluation}

In our experiments, we adopted the Equal Error Rate (EER), and ROC curves with Area Under Curve (AUC) as measures to evaluate and compare the accuracy techniques. Note that small EER and high AUC indicate the good performance of an algorithm.

\section{Experiment results and discussion}
\label{sec:experiment}

In this section, we present the experimental results and analyze on TALKIN videos for different uni-modal and multi-modal kinship verification methods.

\subsection{Uni-modal kinship verification}


\begin{table}[]
\centering
\caption{EER (\%) for the face modality on TALKIN dataset}
\label{tab:face-results-EER}
\newcommand{\tabincell}[2]{\begin{tabular}{@{}#1@{}}#2\end{tabular}}
\begin{tabular}{|l||c|c|c|c|c|}
\hline
\bf{Techniques}     & FS     & FD     & MS     & MD     & \tabincell{c}{Average} \\ \hline \hline
BSIF-Average~\cite{bsif_icpr12}  & 49.0 & 50.0 & 46.0 & 45.0 & 47.5\\ \hline
LPQ-Average~\cite{lpq_isp08} & 44.0 & 50.0 & 50.0 & 50.0   & 48.5\\ \hline
LBP-Average~\cite{ahonen2006face,ojala1996comparative} & 50.0 & 44.0 & 45.0 & 46.0 & 46.3\\ \hline
LBP-TOP~\cite{zhao2007dynamic} & 45.0 & 46.0 & 38.0 & 49.0 & 44.5\\ \hline
VGG-Face + LSTM & \textbf{27.0} & \textbf{35.0} & \textbf{34.0} & \textbf{34.0} & \textbf{32.5} \\ \hline
\end{tabular}
\end{table}


\begin{table}[]
\centering
\caption{EER (\%) for the voice modality on TALKIN dataset.}
\label{Audio results---EER}
\newcommand{\tabincell}[2]{\begin{tabular}{@{}#1@{}}#2\end{tabular}}
\begin{tabular}{|l||c|c|c|c|c|}
\hline
\bf{Techniques} & FS & FD & MS & MD & \tabincell{c}{Average} \\ \hline \hline
I-vector~\cite{dehak2011front} & 47.0 & 47.0 & 44.0 & 48.0 & 46.5     \\ \hline
GMM-UBM~\cite{reynolds_speaker_2000}  & 48.0 & 50.0 & \textbf{44.0} & 45.0 & 46.8    \\ \hline
Resnet-50 & \textbf{32.0} & \textbf{45.0} & \textbf{44.0} & \textbf{40.0}  & \textbf{40.3}\\ \hline
\end{tabular}
\end{table}

\subsubsection{Face-based kinship verification}

Table~\ref{tab:face-results-EER} shows the EERs of visual kinship verification from four relations and average accuracy. From the overall average, our proposed VGG cascaded with a layer of LSTM shows better performance with about more than 29.3\% lower EER compared with hand-crafted features. 

The EERs in Table \ref{tab:face-results-EER} indicate the difficulty of kinship detection from faces. The EERs are notoriously high. In fact, as the chance level is 50\%, the hand-crafted feature extraction techniques shown in the first four rows do little (or no) better than random guessing. This may not be surprising, remembering that none of the methods uses kinship/family labels. Equivalently, cosine scoring assumes all the feature dimensions to be equally informative, which may not hold for in-the-wild data such as TALKIN.
Even if the EERs from the VGG + LSTM approach indicate performance better than random guessing, the error rates are too high to be of practical relevance. This motivates the study of voice-based and bi-modal kinship methods. 


\subsubsection{Voice-based kinship verification}
EERs for voice-based kinship verification are shown in Table
~\ref{Audio results---EER}. 
As with the face modality, the EERs are high. There is little difference between the two GMM-based techniques (GMM-UBM and i-vector), both yielding results close to the chance rate. As expected, here too the deep approach (Resnet-50) provides the lowest overall EER.



\subsubsection{Comparison of face and voice cases}

Tables~\ref{tab:face-results-EER} and ~\ref{Audio results---EER} indicate that deep modal trained with Siamese fashion has the potential to outperform traditional rule-based methods. 

\begin{table}[]
\centering
\caption{EER (\%) from uni-modal and multi-modal techniques on TALKIN dataset.}
\label{results-all-eer}
\newcommand{\tabincell}[2]{\begin{tabular}{@{}#1@{}}#2\end{tabular}}
\begin{tabular}{|l||c|c|c|c|c|}
\hline 
\bf{Techniques} & FS & FD & MS & MD & \tabincell{c}{Average} \\ \hline \hline
Resnet-50 (audio) & 32.0 & 45.0 & 44.0 & 40.0 & 40.1 \\ \hline
VGG+LSTM (video) & 27.0 & 35.0 & 34.0 & 34.0 & 32.5 \\ \hline
Late fusion    & \textbf{20.0} & 37.0 & 37.0  & \textbf{30.0} & 31.0     \\ \hline
Early fusion  & \textbf{20.0} & 38.0 & 35.0 & \textbf{30.0} & 30.8     \\ \hline
\tabincell{c}{Deep Siamese\\ Network (ours)} & 23.0 & \textbf{34.0} & \textbf{31.0} & 31.0 & \textbf{29.8}     \\ \hline
\end{tabular}
\end{table}

\begin{figure}[htb!]
\centering
\subfigure[FS relation]{\includegraphics[width=1.67in]{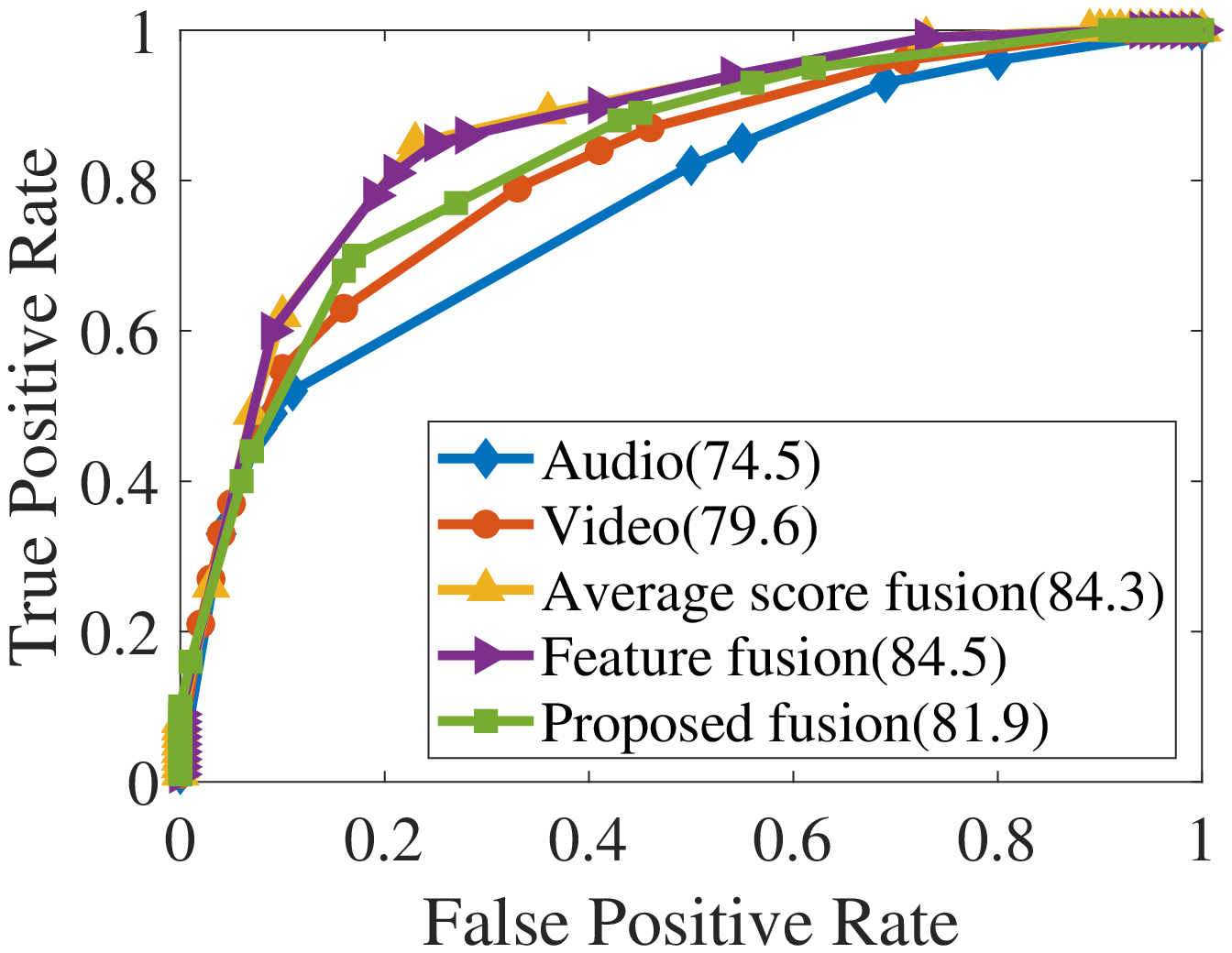}
\label{FS}}
\hfil
\subfigure[FD relation]{\includegraphics[width=1.67in]{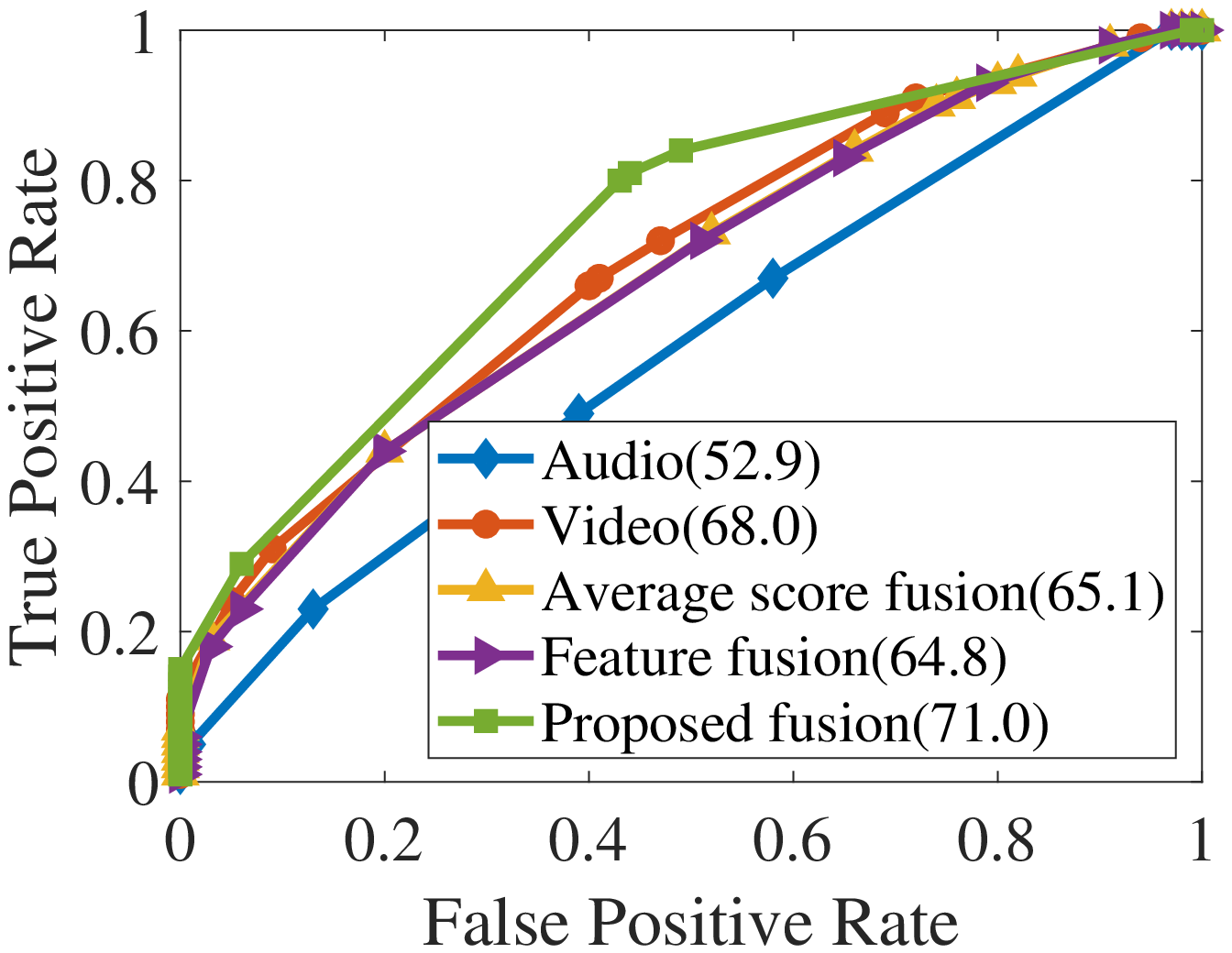}
\label{FD}}
\hfil
\subfigure[MS relation]{\includegraphics[width=1.67in]{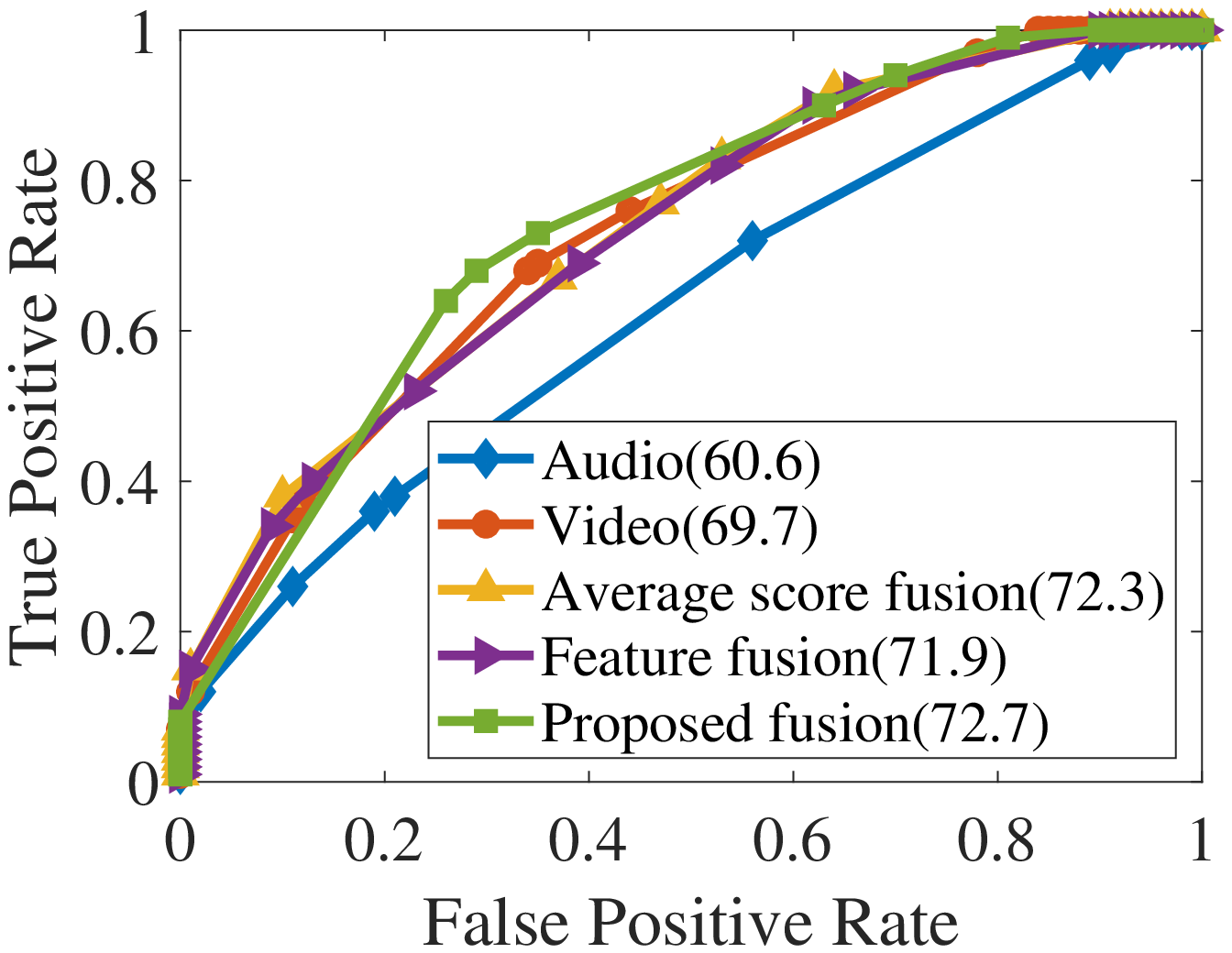}
\label{MS}}
\hfil
\subfigure[MD relation]{\includegraphics[width=1.67in]{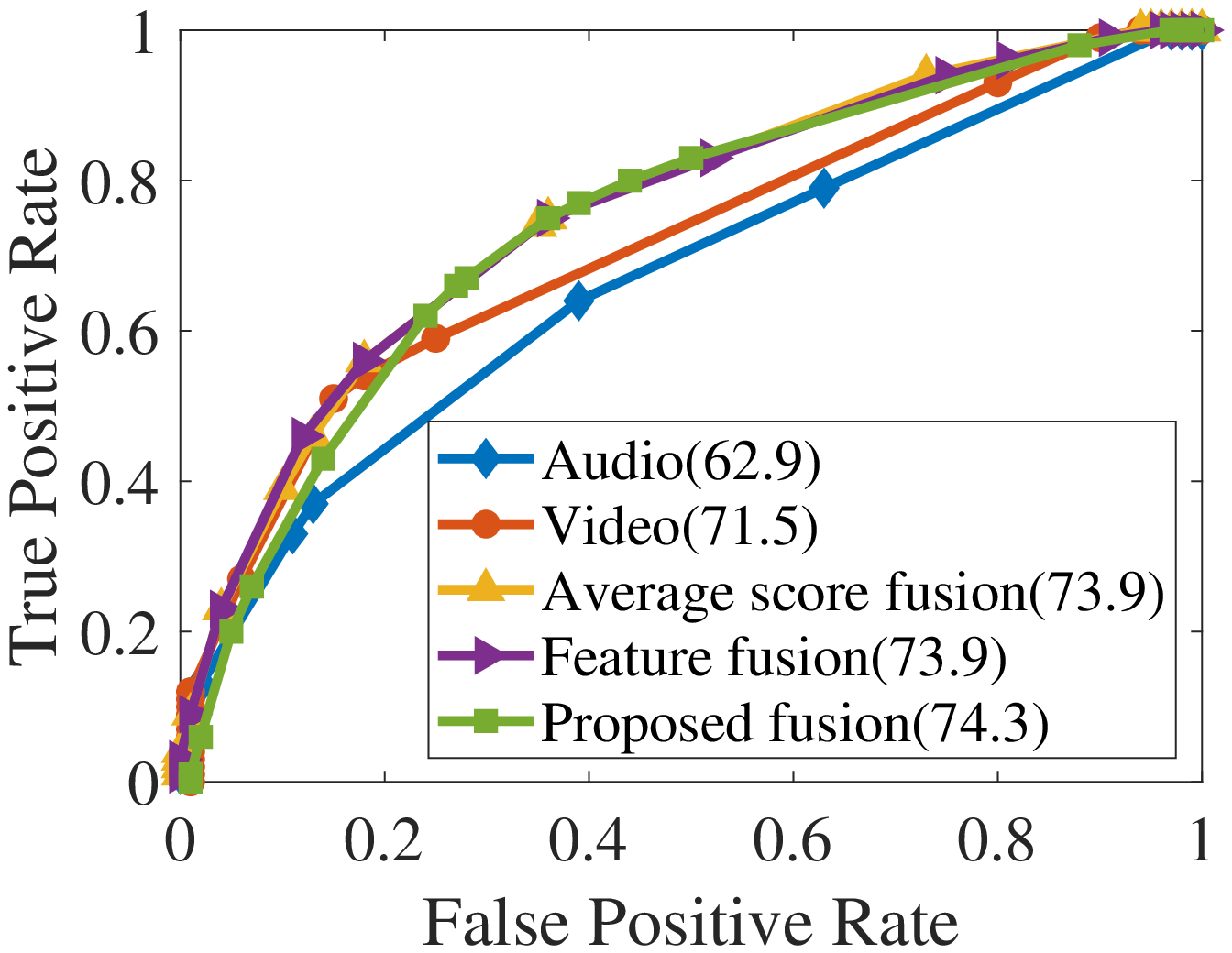}
\label{MD}}
\hfil
\caption{ROC curves uni- and multi-modal techniques for kinship verification on TALKIN dataset. The numbers in parentheses are the Area Under the ROC Curve for each method.}
\label{fig: ROC_fusion}
\end{figure}

\subsection{Multi-Modal Kinship Verification}
\label{subsec:multi-modal}

The comparison of different fusion methods and uni-modal performance is given in Table~\ref{results-all-eer} for EERs and Fig.~\ref{fig: ROC_fusion} for ROC curves. In Fig.~\ref{fig: ROC_fusion}, \emph{area under the ROC curve} (AUC) values are also provided. From Table~\ref{results-all-eer}, proposed fusion method with deep Siamese network gets the highest performance in average EER. For FS and MD relation, early fusion and late fusion get the best performance with EER of 20.0\% and 30.0\% separately.

Compared with uni-modal kinship verification methods, fusing both face and voice modalities can lead to better performance with about 3\%-10\% lower in average EER, which also demonstrates that face and voice modalities can give complementary information in kinship verification task. Multi-modal techniques can help to improve the robustness of kinship verification system.

\section{Conclusion and Future Directions}
\label{sec:conclusion}

Using machine learning techniques for kinship verification has become a application of interest within the computer vision committee. 
Inspired by a study where people with a kinship relation share similar vocal features and can confuse the speaker verification system, we proposed leveraging vocal information for kinship verification.

In the absence of a kinship database that contains vocal information, we collected a new TALKIN kinship database that is comprised of both facial and vocal information captures from videos while subjects talking. First, we conducted experiments for uni-modal kinship verification from both face and voice aspects. Two state-of-the-art deep architectures (face \& voice) were trained in a Siamese fashion with contrastive loss to provide the best average accuracy. We also proposed a deep Siamese fusion network for kinship verification to combine visual and vocal information that compares favourably to baseline late and early fusion methods. The experimental results also showed that multi-modal kinship verification provide a higher level of accuracy compared with uni-modal kinship verification.

In the future, we plan to investigate deep architectures for spatio-temporal fusion of visual and audio signals. Discriminative analysis will be carried out to explore the discriminative capability of face modality and voice modalities. Additionally, the efficiency of deep learning models for feature extraction and fusion are also a concern. We are currently enlarging the database and planning to make it publicly available for the research community.

\section*{Acknowledgment}
The authors wish to acknowledge CSC-IT Center for Science, Finland, for computational resources. The initial help from Dr. Miguel Bordallo López and Dr. Elhocine Boutellaa is also acknowledged. The authors wish to thank T.H.~Kinnunen and A. Hadid for their technical advice for this paper.

\ifCLASSOPTIONcaptionsoff
  \newpage
\fi
\bibliographystyle{IEEEtran}
\bibliography{reference}

\begin{IEEEbiography}[{\includegraphics[width=1in,height=1.25in,clip,keepaspectratio]{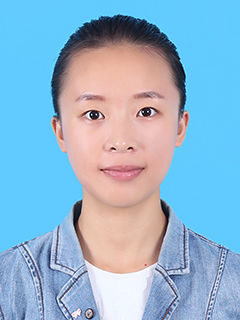}}]{Xiaoting Wu}
is currently working toward the Ph.D. degree in the Center for Machine Vision and Signal Analysis, University of Oulu, Oulu, Finland and School of Electronics and Information, Northwestern Polytechnical University, Xi'an, China. She received her Bachelor degree in Lanzhou University, China in 2014 and Master degree in Northwestern Polytechnical University, China in 2016. Her research interests include machine vision, machine learning and kinship verification.
\end{IEEEbiography}


%
\begin{IEEEbiography}
[{\includegraphics[width=1in,height=1.25in,clip,keepaspectratio]{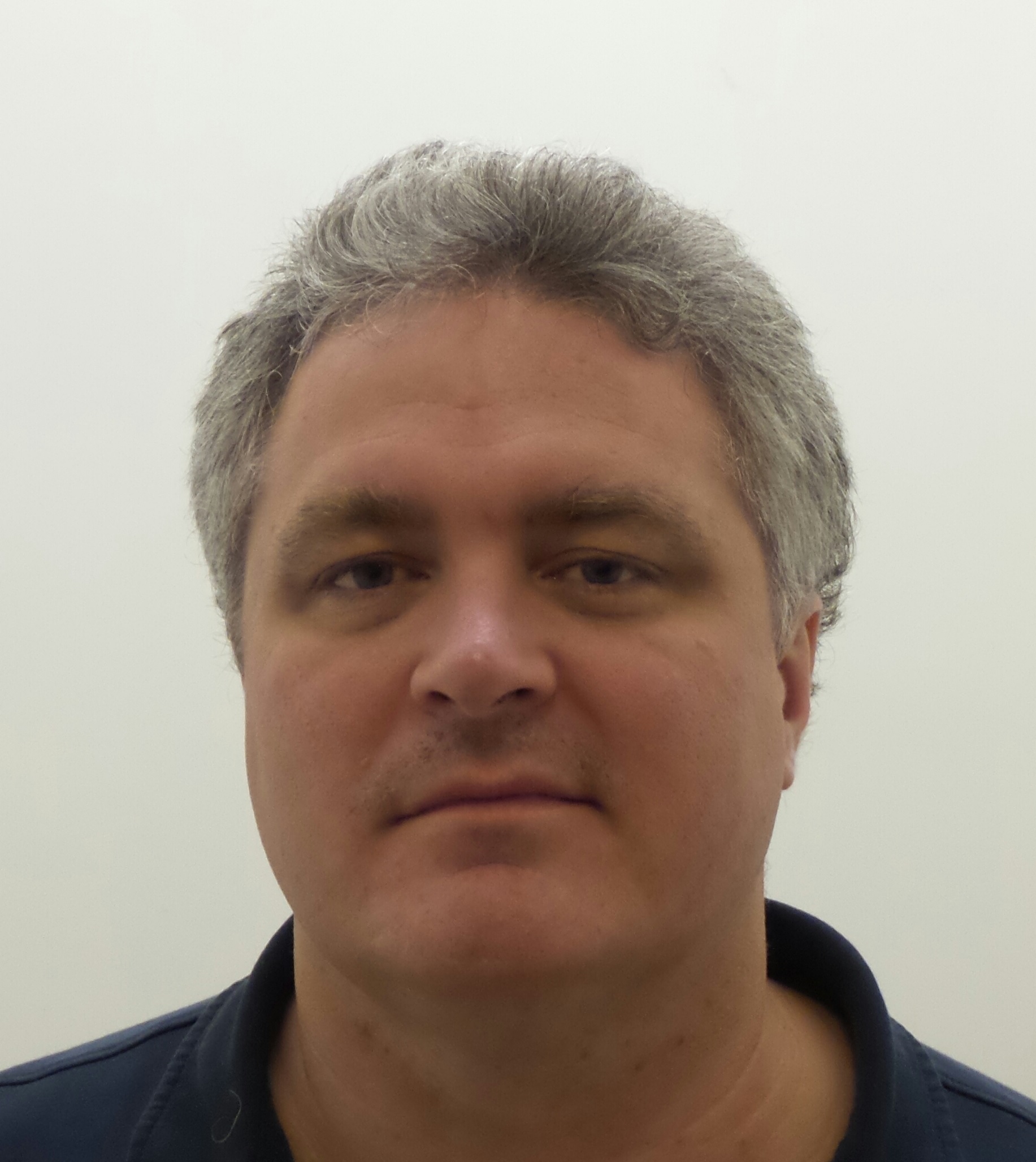}}]{Eric~Granger}
 (M'00) received Ph.D. in Electrical Engineering from \'Ecole Polytechnique de Montr\'eal in 2001, and worked as a Defense Scientist at DRDC-Ottawa (1999-2001), and in R\&D with Mitel Networks (2001-04). In 2004, he joined the \'Ecole de technologie sup\'erieure (Universit\'e du Qu\'ebec), Montreal, where he is currently a Full Professor and the Director of Laboratoire d'imagerie, de vision et d'intelligence artificielle, a research laboratory focused on computer vision and artificial intelligence.  His research interests include pattern recognition, machine learning, computer vision, domain adaptation, and incremental and weakly-supervised learning, with applications in biometrics, affective computing, video surveillance, and computer/network security.
\end{IEEEbiography}

\begin{IEEEbiography}
[{\includegraphics[width=1in,height=1.25in,clip,keepaspectratio]{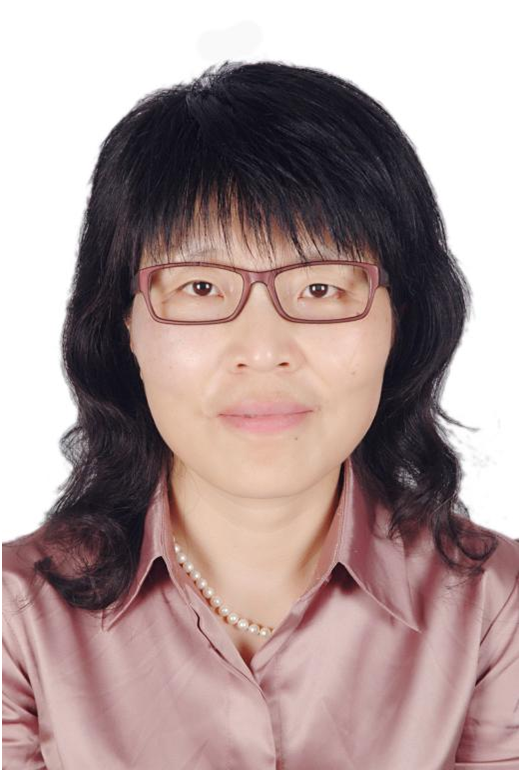}}]{Xiaoyi~Feng}
received the M.S. degree from the Northwest University, Xi'an, China, in 1994. She received her Ph.D. degree from the Northwestern Polytechnical University, Xi'an, China, in 2001. She is currently a professor with the School of Electronics and Information, Northwestern Polytechnical University since 2008. Her current research interests include computer vision, image process, radar imagery and recognition.
\end{IEEEbiography}


\end{document}